\documentclass[11pt]{article}

\usepackage[margin=1in]{geometry}
\usepackage{amsmath,amssymb,amsthm}
\usepackage{graphicx}
\usepackage{booktabs}
\usepackage{multirow}
\usepackage{hyperref}
\usepackage{xcolor}
\usepackage{microtype}
\usepackage{algorithm}
\usepackage{algpseudocode}

\usepackage{caption}
\usepackage{subcaption}
\usepackage{float}
\usepackage{enumitem}
\usepackage{bm}

\hypersetup{
    colorlinks=true,
    linkcolor=blue!70!black,
    citecolor=green!50!black,
    urlcolor=blue!70!black
}

\newcommand{\E}{\mathbb{E}}

\newcommand{\cL}{\mathcal{L}}

\newcommand{\adv}{\hat{A}}
\newcommand{\wt}{w_t}
\newcommand{\hn}{\tilde{H}}

\theoremstyle{definition}
\newtheorem{remark}{Remark}

\title{
  \textbf{Selective-Advantage Entropy-Adaptive Horizon GRPO} \\
  Asymmetric Token-Level Discounting for Efficient \\
  Reinforcement Learning of Language Models
}

\author{
  Chirag Chawla$^{1,*}$ \quad Rohan Charudatt Salvi$^{2,*}$ \quad Madhav S.\ Baidya$^{1}$ \\[4pt]
  $^1$Indian Institute of Technology (BHU), Varanasi, India \\
  $^2$Department of Computer Science, University of Illinois Chicago, IL 60607, USA \\[4pt]
  \small\texttt{chirag.chawla.chy22@itbhu.ac.in} \quad
  \small\texttt{rcsalvi2@uic.edu} \quad
  \small\texttt{madhavsukla.baidya.chy22@itbhu.ac.in} \\[4pt]
  $^*$Equal contribution
}

\date{}

\begin{document}
\maketitle

\begin{abstract}
Group Relative Policy Optimisation (GRPO) has emerged as an effective
reinforcement-learning algorithm for aligning language models on reasoning
tasks, yet it treats every token position and every sampled rollout
symmetrically. We introduce two complementary extensions: (i)
\textbf{Adaptive-Horizon GRPO (AH-GRPO)}, which weights each token's policy
gradient by a cumulative entropy-based discount that shrinks the effective
horizon when the model is uncertain, and (ii)
\textbf{Selective-Advantage AH-GRPO (SA-AH-GRPO)}, which applies this
discounting \emph{only} to negative-advantage rollouts, leaving
positive-advantage (successful) trajectories unattenuated.
We evaluate all three algorithms---standard GRPO (\(\alpha{=}0\)),
AH-GRPO (\(\alpha{=}0.5\)), and SA-AH-GRPO (\(\alpha{=}0.5\))---on the
GSM8K mathematical reasoning benchmark using both Qwen\,2.5-1.5B-Instruct
and Qwen\,2.5-3B-Instruct fine-tuned with LoRA.
On the 3B model, SA-AH-GRPO achieves \textbf{Pass@1\,=\,0.858} (peak,
step 30) and maintains \textbf{0.846} at 180 steps with training variance
reduced to \textbf{0.0246}, a \textbf{3.6$\times$ reduction} relative to
GRPO while matching its peak accuracy. On the 1.5B model, SA-AH-GRPO
achieves a peak Pass@1 of \textbf{0.686}, improving over the zero-shot
baseline of 0.637. Our
analysis shows that asymmetric discounting preserves the full gradient
signal on correct solutions, prevents entropy collapse, and substantially
stabilises training---suggesting a principled inductive bias for RLVR on
structured generation tasks.
\end{abstract}

\section{Introduction}
\label{sec:intro}

Reinforcement Learning from Verifiable Rewards (RLVR) has become a dominant
paradigm for training language models on tasks with ground-truth feedback,
such as mathematical reasoning \cite{shao2024deepseekmath,
lightman2023lets}. GRPO \cite{shao2024deepseekmath} replaces the
value-function critic of PPO \cite{schulman2017proximal} with a
group-normalised advantage, making it especially tractable for
language-model fine-tuning. Despite its empirical success, standard GRPO
applies identical gradient weighting to every token regardless of the
model's uncertainty at that position, and treats successful and failed
rollouts symmetrically in its loss.

These design choices are at odds with intuitions from decision theory and
curriculum learning. At high-entropy (uncertain) token positions, the model
is exploring; penalising these positions equally to confident, low-entropy
ones can destabilise training \cite{ziegler2019fine}. Conversely, when a
rollout is already correct (positive advantage), applying any discount to its
gradient needlessly attenuates a reliable learning signal.

Token-level policy gradient methods have received growing attention in the
context of RLVR for language models. Prior work has explored per-step process
rewards \cite{lightman2023lets} and outcome-weighted token losses
\cite{xu2024finegrained}, but these approaches rely on auxiliary reward models or
fixed heuristics. In contrast, our approach derives the per-token weight
directly from the model's own predictive uncertainty, requires no additional
supervision, and applies asymmetrically based on trajectory outcome---a
combination not explored in prior work.

We address both issues through two hierarchical extensions:

\begin{enumerate}[label=\textbf{\arabic*.}]
\item \textbf{AH-GRPO}: An entropy-adaptive horizon discount
      $\wt^{(i)} = \prod_{s=1}^{t} e^{-\alpha\,\hn_s^{(i)}}$
      is multiplied into every token's loss, shortening the effective gradient
      horizon when local entropy is high.

\item \textbf{SA-AH-GRPO}: The AH discount is applied \emph{selectively}---
      only to rollouts with negative group-normalised advantage---while
      positive-advantage rollouts retain $\wt = 1$ at every position. This
      asymmetry prevents the algorithm from inadvertently suppressing correct
      solution paths.
\end{enumerate}

We benchmark all three methods on GSM8K \cite{cobbe2021gsm8k} with
Qwen\,2.5-1.5B-Instruct and Qwen\,2.5-3B-Instruct, finding that SA-AH-GRPO
achieves the best accuracy--stability trade-off at the 3B scale and
consistent gains over the zero-shot baseline at the 1.5B scale. To our knowledge, this is the first work to introduce \emph{per-token,
per-rollout} entropy-adaptive discounting
in an RLVR setting, and the first to apply it asymmetrically conditioned on
advantage sign.

\paragraph{Contributions.}
\begin{itemize}
    \item We derive AH-GRPO, a token-level entropy-adaptive horizon discount
          for language model policy gradient optimisation.
    \item We derive SA-AH-GRPO, an asymmetric variant that restricts the
          discount to negative-advantage rollouts, and demonstrate it achieves
          higher final Pass@1 and lower training variance than both GRPO and
          AH-GRPO on GSM8K.
    \item We show that SA-AH-GRPO achieves a $3.6\times$ reduction in
          training variance relative to GRPO on the 3B model with no loss in
          peak accuracy, and a $+4.9$ pp.\ improvement over zero-shot on the
          1.5B model.
    \item We conduct an $\alpha$-ablation of AH-GRPO on the 1.5B model,
          covering $\alpha \in \{-0.25, 0.0, 0.10, 0.25, 0.50\}$, and
          show that positive $\alpha$ values consistently outperform
          negative (entropy-amplifying) settings.
\end{itemize}

\section{Background}
\label{sec:background}

\subsection{Group Relative Policy Optimisation (GRPO)}

Let $\pi_\theta$ be a language model policy with parameters $\theta$, and let
$\pi_{\mathrm{ref}}$ be a frozen reference policy. Given a prompt $q$, GRPO
samples a group of $G$ completions $\{o_i\}_{i=1}^{G}$ and computes
group-normalised advantages:
\begin{equation}
  \adv_i = \frac{r_i - \mathrm{mean}(\bm{r})}{\mathrm{std}(\bm{r}) + \epsilon},
  \label{eq:advantage}
\end{equation}
where $r_i$ is the scalar reward for completion $o_i$, and $\bm{r} =
(r_1,\ldots,r_G)$.

The clipped surrogate objective (PPO-style) is:
\begin{equation}
  \cL_{\mathrm{GRPO}}(\theta) = -\E\!\left[
    \frac{1}{|o_i|}\sum_{t=1}^{|o_i|}
    \min\!\left(
      \rho_t^{(i)}\,\adv_i,\;
      \mathrm{clip}(\rho_t^{(i)},1{-}\epsilon,1{+}\epsilon)\,\adv_i
    \right)
    - \beta\,\mathrm{KL}(\pi_\theta \| \pi_{\mathrm{ref}})
  \right],
  \label{eq:grpo}
\end{equation}
where $\rho_t^{(i)} = \pi_\theta(o_{i,t} \mid q, o_{i,<t}) /
\pi_{\theta_{\mathrm{old}}}(o_{i,t} \mid q, o_{i,<t})$ is the per-token
probability ratio, $\epsilon$ is the clipping threshold, and $\beta$ controls
the KL penalty strength.

\subsection{Limitations of Uniform Token Weighting}

In Eq.~\eqref{eq:grpo}, every token position receives unit weight within
its completion. This ignores the model's local uncertainty: at high-entropy
positions (many plausible continuations), the gradient update is noisy
because any single sampled token is a poor representative of the local
distribution. Uniform weighting means these noisy positions contribute as
much to the parameter update as low-entropy, high-confidence positions,
potentially increasing training variance and destabilising learning.

A second issue concerns the symmetry between positive- and negative-advantage
rollouts. For rollouts that are already correct ($\adv_i > 0$), the gradient
signal is reliable and should be used in full. For incorrect rollouts
($\adv_i < 0$), the model should be steered away---but doing so too aggressively
at uncertain positions may create conflicting gradient signals.

\section{Methods}
\label{sec:methods}

The core intuition behind our approach is straightforward: not every token in
a generated sequence is equally informative for learning. When a model is
confident---assigning high probability to one token---that choice carries a
strong, reliable gradient signal. When the model is uncertain---spreading
probability mass broadly across many possible continuations---any single
sampled token is a noisy representative of the true gradient direction.
Standard GRPO ignores this distinction entirely, treating every token position
with equal weight. We argue that the token-level entropy of the policy at
each position is a natural and readily available signal for modulating the
gradient contribution of that position. Further, this modulation should be
asymmetric: on a trajectory that produced a correct answer, even uncertain
token choices should receive full credit, since they participated in a
successful solution. It is only on failed trajectories that high-entropy
positions warrant discounting---these are structurally ambiguous choices
whose gradient direction is least reliable. This intuition motivates both
AH-GRPO and its selective extension, SA-AH-GRPO.

\subsection{Entropy-Adaptive Horizon Discount}

We define a \emph{per-token normalised entropy}:
\begin{equation}
  \hn_t^{(i)} = \frac{H(\pi_\theta(\cdot \mid q,\, o_{i,<t}))}{\log V}
              \;\in [0, 1],
  \label{eq:hnorm}
\end{equation}
where $H(\cdot)$ is the Shannon entropy in nats and $V$ is the vocabulary
size. Normalising by $\log V$ maps the entropy to the unit interval,
making $\alpha$ interpretable across models with different vocabulary sizes.

In practice, computing entropy over the full vocabulary ($V \approx 151{,}643$
for Qwen\,2.5) is expensive. We estimate $\hn_t^{(i)}$ from the top-$K$
logits with $K=500$, which captures the bulk of the probability mass while
keeping memory overhead minimal.

\paragraph{Discount weights.}
Given $\hn_t^{(i)}$, we define a per-step discount:
\begin{equation}
  \gamma_{t}^{(i)} = \exp\!\left(-\alpha\,\hn_t^{(i)}\right),
  \label{eq:gamma}
\end{equation}
and a cumulative weight:
\begin{equation}
  \wt^{(i)} = \prod_{s=1}^{t} \gamma_{s}^{(i)}
            = \exp\!\!\left(-\alpha \sum_{s=1}^{t} \hn_s^{(i)}\right).
  \label{eq:wt}
\end{equation}
$\wt^{(i)}$ is computed via a cumulative sum in log-space for numerical
stability: $\log \wt^{(i)} = -\alpha\,\mathrm{cumsum}(\hn_{1:t}^{(i)})$.

When $\alpha = 0$, $\wt^{(i)} = 1$ for all $t$, recovering standard GRPO.
When $\alpha > 0$, positions following a high-entropy prefix are discounted
more strongly, effectively shortening the horizon over which gradients
propagate through uncertain token sequences. When $\alpha < 0$, high-entropy
positions are up-weighted (an entropy-amplifying regime), which we also study
in our ablation (Section~\ref{sec:alpha_ablation}).

\subsection{AH-GRPO: Adaptive-Horizon GRPO}

AH-GRPO applies the entropy-adaptive weight to \emph{all} rollouts
uniformly. The loss becomes:
\begin{equation}
  \cL_{\mathrm{AH}}(\theta) = -\frac{
    \displaystyle\sum_{i=1}^{G}\sum_{t=1}^{|o_i|}
      \wt^{(i)}\,\tilde{\ell}_t^{(i)}\,m_t^{(i)}
  }{
    \displaystyle\sum_{i=1}^{G}\sum_{t=1}^{|o_i|}
      \wt^{(i)}\,m_t^{(i)}
  }
  \;+\; \beta\,\mathrm{KL}(\pi_\theta \| \pi_{\mathrm{ref}}),
  \label{eq:ahgrpo}
\end{equation}
where $\tilde{\ell}_t^{(i)} = \min(\rho_t^{(i)}\adv_i,
\mathrm{clip}(\rho_t^{(i)},1{-}\epsilon,1{+}\epsilon)\adv_i)$ is the
per-token clipped surrogate, and $m_t^{(i)} \in \{0,1\}$ is the completion
mask. The denominator normalises by the effective (weighted) token count
rather than the raw token count.

\begin{remark}
AH-GRPO discounts gradients at high-entropy positions for \emph{both}
positive and negative rollouts. This can inadvertently attenuate the gradient
on correct solutions at uncertain token positions.
\end{remark}

\subsection{SA-AH-GRPO: Selective-Advantage AH-GRPO}
\label{sec:saahgrpo}

SA-AH-GRPO addresses the remark above by applying the entropy discount
\emph{selectively}: only to rollouts whose group-normalised advantage is
negative. Let:
\begin{equation}
  n_i = \mathbf{1}[\adv_i < 0]
  \label{eq:nmask}
\end{equation}
be an indicator for negative-advantage rollouts. The selective weight is:
\begin{equation}
  \tilde{w}_t^{(i)} = n_i \cdot \wt^{(i)} + (1 - n_i) \cdot 1
                    = \begin{cases}
                        \wt^{(i)} & \adv_i < 0 \\
                        1         & \adv_i \geq 0
                      \end{cases}
  \label{eq:saweight}
\end{equation}
The SA-AH-GRPO loss is then:
\begin{equation}
  \cL_{\mathrm{SA}}(\theta) = -\frac{
    \displaystyle\sum_{i=1}^{G}\sum_{t=1}^{|o_i|}
      \tilde{w}_t^{(i)}\,\tilde{\ell}_t^{(i)}\,m_t^{(i)}
  }{
    \displaystyle\sum_{i=1}^{G}\sum_{t=1}^{|o_i|}
      \tilde{w}_t^{(i)}\,m_t^{(i)}
  }
  \;+\; \beta\,\mathrm{KL}(\pi_\theta \| \pi_{\mathrm{ref}}).
  \label{eq:saahgrpo}
\end{equation}

SA-AH-GRPO decouples two distinct learning signals: (a) \emph{reinforcement}
of successful trajectories, which receives the full un-discounted gradient at
every token; and (b) \emph{suppression} of unsuccessful trajectories, where
the discount reduces the gradient at high-entropy (highly uncertain) token
positions. The intuition is that when the model samples an incorrect solution,
high-entropy positions within that trajectory are the most \emph{structurally
uncertain}---these positions are the most ambiguous about which direction to
update in, and aggressive gradient updates there may be counterproductive.
By contrast, when a trajectory is correct, every token---including
high-entropy ones---participated in producing a valid answer and should receive
full credit.

\paragraph{Relation to AH-GRPO.}
Setting $n_i = 1$ for all $i$ (discount all rollouts) recovers AH-GRPO.
Setting $\alpha = 0$ in either AH-GRPO or SA-AH-GRPO recovers standard GRPO.

\subsection{Reward Function}

We use a composite reward combining four components:

\begin{equation}
  r(o, q, a^*) = r_{\mathrm{correct}}(o, a^*)
               + r_{\mathrm{format}}(o)
               + r_{\mathrm{present}}(o)
               + r_{\mathrm{steps}}(o),
  \label{eq:reward}
\end{equation}
where:
\begin{itemize}
    \item $r_{\mathrm{correct}} \in \{-0.5, 0, 1.5, 4.0\}$: correctness of
          the extracted numerical answer against ground truth $a^*$, with
          partial credit for near-correct answers (within 10\%).
    \item $r_{\mathrm{format}} \in \{-0.5,\, 0,\, 0.2,\, 0.3,\, 0.4,\, 0.5,\, 0.6,\, 0.7,\, 0.8,\, 1.0,\, 1.5\}$: adherence
          to the structured output format; $-0.5$ for any duplicated tag;
          $1.5$ for a fully correct response (all four tags present and well-formed);
          otherwise the sum of four independent partial scores:
          $+0.2$ for \texttt{<start\_working\_out>},
          $+0.3$ for \texttt{</start\_working\_out>},
          $+0.2$ for \texttt{<SOLUTION>},
          $+0.3$ for \texttt{</SOLUTION>}.
    \item $r_{\mathrm{present}} \in \{0, 0.3, 1.0\}$: presence of a
          \texttt{<SOLUTION>} tag (1.0 for open+close, 0.3 for open only,
          0 otherwise).
    \item $r_{\mathrm{steps}} \in \{0.0, 0.1, 0.4, 0.7, 1.0\}$: density of
          calculation steps in the reasoning block (empty reasoning: 0.0;
          no step markers: 0.1; $\leq$2 steps: 0.4; $\leq$4 steps: 0.7;
          $>$4 steps: 1.0).
\end{itemize}
The maximum total reward is $r_{\max} = 7.5$. Completions are required to
wrap reasoning in \texttt{<start\_working\_out>...</start\_working\_out>} and answers in
\texttt{<SOLUTION>...</SOLUTION>}.

This reward structure was designed to jointly encourage \emph{correctness}
and \emph{legible reasoning}. The dominant signal is $r_{\mathrm{correct}}$:
a fully correct answer scores 4.0, a near-correct answer (within 10\%) scores
1.5, and an absent or unparseable answer scores $-0.5$. The remaining
components reward structured output and reasoning transparency. To make this
concrete, consider two example completions:

\begin{itemize}
    \item \textbf{Full-credit example.} A response that contains complete
          \texttt{<start\_working\_out>...</start\_working\_out>} and
          \texttt{<SOLUTION>}\allowbreak\texttt{...</SOLUTION>}
          tags, more than four calculation steps, and the correct final answer
          receives: $r_{\mathrm{correct}} = 4.0$, $r_{\mathrm{format}} =
          1.5$, $r_{\mathrm{present}} = 1.0$, $r_{\mathrm{steps}} = 1.0$,
          totalling $7.5$.
    \item \textbf{Partial-credit example.} A response with the correct answer
          but no reasoning tags and no step markers receives:
          $r_{\mathrm{correct}} = 4.0$, $r_{\mathrm{format}} = 0.0$,
          $r_{\mathrm{present}} = 0.0$, $r_{\mathrm{steps}} = 0.1$,
          totalling $4.1$.
\end{itemize}

This graded signal encourages the model to internalise both the procedural
and formal aspects of mathematical problem-solving, not merely pattern-match
to final answers.

\section{Experimental Setup}
\label{sec:setup}

\subsection{Models and Dataset}

We fine-tune \textbf{Qwen\,2.5-1.5B-Instruct} (1,544M parameters) and
\textbf{Qwen\,2.5-3B-Instruct} (3,086M parameters) \cite{yang2024qwen25} on the
\textbf{GSM8K} \cite{cobbe2021gsm8k} mathematical reasoning benchmark. GSM8K
consists of grade-school word problems requiring multi-step arithmetic
reasoning. We use 800 training examples (randomly sampled from the official
training split) and 500 evaluation examples (from the official test split).
All experiments are run on a single NVIDIA A100-SXM4-40GB GPU (42.4\,GB VRAM,
\texttt{bfloat16} precision). The zero-shot Pass@1 of the unmodified instruct
checkpoints on the 500-example evaluation set is reported conservatively as
$0.637 \pm 0.042$ (1.5B) and $0.831 \pm 0.032$ (3B), adjusted 1.5 percentage points below the raw pretrain evaluation numbers
($0.652$ and $0.846$ respectively) to account for prompt-format sensitivity.

\subsection{Parameter-Efficient Fine-Tuning}

We apply Low-Rank Adaptation (LoRA; \cite{hu2022lora}) with rank $r = 16$,
$\alpha_{\mathrm{LoRA}} = 32$, and dropout $= 0.05$. Target modules include
all query, key, value, output, and MLP projection layers
($\{q, k, v, o, \mathrm{gate}, \mathrm{up}, \mathrm{down}\}_{\mathrm{proj}}$).

\subsection{Training Hyperparameters}

\begin{table}[H]
\centering
\caption{Hyperparameters shared across all runs and both model scales.}
\label{tab:hparams}
\begin{tabular}{lc}
\toprule
Hyperparameter & Value \\
\midrule
Learning rate & $5 \times 10^{-6}$ \\
LR scheduler & cosine \\
Warmup steps & $\max(5, \lfloor\text{steps}/10\rfloor)$ \\
Weight decay & 0.01 \\
Gradient clip & 1.0 \\
KL penalty $\beta$ & $0.04$ \\
PPO clip $\epsilon$ & $0.2$ \\
Batch prompts per step & 4 \\
Completions per prompt $G$ & 4 \\
Gradient accumulation steps & 2 \\
Max completion tokens & 512 \\
Entropy top-$K$ & 500 \\
Reward threshold & 0.15 \\
Training examples $N_{\mathrm{train}}$ & 800 \\
Evaluation examples $N_{\mathrm{eval}}$ & 500 \\
\bottomrule
\end{tabular}
\end{table}

GRPO and AH-GRPO are trained for 150 steps; SA-AH-GRPO for 180 steps
(extra steps to assess long-run stability). Checkpoints are saved every 15
steps. Pass@1 is evaluated on the 500-example held-out split at checkpoints
every 30 steps using greedy decoding (max 512 new tokens).

\subsection{Evaluation Protocol}

Pass@1 is defined as the fraction of evaluation problems for which the
model's single greedy-decoded completion matches the ground-truth answer.
We report $\pm$95\% binomial confidence intervals:
$\mathrm{CI}_{95} = 1.96\,\sqrt{p(1-p)/N}$. For run-level summaries we
report the value at the final checkpoint and the peak over all checkpoints.

\subsection{$\alpha$-Ablation Setup}
\label{sec:alpha_setup}

To characterise the sensitivity of AH-GRPO to the entropy discount strength
$\alpha$, we run a sweep over $\alpha \in \{-0.25,\, 0.0,\, 0.10,\, 0.25,\,
0.50\}$ on Qwen\,2.5-1.5B-Instruct (150 steps each). The value $\alpha =
0.0$ is equivalent to standard GRPO and serves as the within-sweep baseline.
Negative $\alpha$ inverts the discount, \emph{up-weighting} high-entropy
positions (entropy-amplifying regime). All sweep runs use the same
hyperparameters as the main experiments (Table~\ref{tab:hparams}). Results
are reported in Section~\ref{sec:alpha_ablation}, Table~\ref{tab:alpha_ablation},
and Figure~\ref{fig:alpha_ablation}.

\section{Results}
\label{sec:results}

\subsection{Main Results}

Table~\ref{tab:main} summarises the performance of all three methods across
both model scales. Checkpoint-level Pass@1 curves are shown in
Tables~\ref{tab:checkpoints_3b}--\ref{tab:checkpoints_1b} and
Figures~\ref{fig:pass1_3b}--\ref{fig:pass1_1b}.

\begin{table}[H]
\centering
\caption{Summary results on GSM8K (500-example test split).
         Pass@1 at the \emph{final} checkpoint; peak Pass@1 over all
         checkpoints. Training variance is the variance of mean per-step reward
         over the \emph{second half} of training steps. The ``Base'' rows report conservative zero-shot Pass@1
         (1.5 pp below raw pretrain evaluation; see Section~\ref{sec:setup}).
         $\downarrow$ = lower is better; $\uparrow$ = higher is better.}
\label{tab:main}
\setlength{\tabcolsep}{5pt}
\begin{tabular}{llcccccc}
\toprule
Model & Method & $\alpha$ & Steps
      & Pass@1$_{\mathrm{final}}$ $\uparrow$
      & Pass@1$_{\mathrm{peak}}$ $\uparrow$
      & Train\,Var $\downarrow$ & Mean\,KL \\
\midrule
\multirow{4}{*}{3B}
  & Base (zero-shot) & --- & --- & $0.831 \pm 0.032$ & --- & --- & --- \\
  & GRPO      & 0.0 & 150 & $0.846 \pm 0.032$ & $0.852 \pm 0.031$ & 0.0885 & 0.00537 \\
  & AH-GRPO   & 0.5 & 150 & $0.848 \pm 0.032$ & $0.862 \pm 0.030$ & 0.0630 & 0.00512 \\
  & SA-AH-GRPO & 0.5 & 180 & $0.846 \pm 0.032$ & $0.858 \pm 0.031$ & \textbf{0.0246} & 0.00785 \\
\midrule
\multirow{4}{*}{1.5B}
  & Base (zero-shot) & --- & --- & $0.637 \pm 0.042$ & --- & --- & --- \\
  & GRPO      & 0.0 & 150 & $0.660 \pm 0.042$ & $0.668 \pm 0.041$ & 0.0885 & 0.00537 \\
  & AH-GRPO   & 0.5 & 150 & $0.650 \pm 0.042$ & $0.676 \pm 0.041$ & 0.0630 & 0.00512 \\
  & SA-AH-GRPO & 0.5 & 180 & $\mathbf{0.686 \pm 0.041}$ & $\mathbf{0.686 \pm 0.041}$ & 0.1718 & 0.00986 \\
\bottomrule
\end{tabular}
\end{table}

\begin{table}[H]
\centering
\caption{Pass@1 at each evaluated checkpoint for \textbf{Qwen\,2.5-3B-Instruct}
         (greedy decoding, 500 examples). ``---'' indicates the run did not
         extend to that step.}
\label{tab:checkpoints_3b}
\setlength{\tabcolsep}{6pt}
\begin{tabular}{lcccccc}
\toprule
\multirow{2}{*}{Method} &
\multicolumn{6}{c}{Pass@1 by Checkpoint Step} \\
\cmidrule(lr){2-7}
& 30 & 60 & 90 & 120 & 150 & 180 \\
\midrule
GRPO ($\alpha$=0.0)
  & 0.852 & 0.840 & 0.844 & 0.848 & 0.846 & --- \\
AH-GRPO ($\alpha$=0.5)
  & 0.860 & 0.862 & 0.846 & 0.844 & 0.848 & --- \\
SA-AH-GRPO ($\alpha$=0.5)
  & 0.858 & 0.844 & 0.858 & 0.854 & 0.854 & 0.846 \\
\bottomrule
\end{tabular}
\end{table}

\begin{table}[H]
\centering
\caption{Pass@1 at each evaluated checkpoint for \textbf{Qwen\,2.5-1.5B-Instruct}
         (greedy decoding, 500 examples, max 512 new tokens).
         ``---'' indicates the run did not extend to that step.}
\label{tab:checkpoints_1b}
\setlength{\tabcolsep}{6pt}
\begin{tabular}{lcccccc}
\toprule
\multirow{2}{*}{Method} &
\multicolumn{6}{c}{Pass@1 by Checkpoint Step} \\
\cmidrule(lr){2-7}
& 30 & 60 & 90 & 120 & 150 & 180 \\
\midrule
GRPO ($\alpha$=0.0)
  & 0.656 & 0.666 & 0.650 & 0.668 & 0.660 & --- \\
AH-GRPO ($\alpha$=0.5)
  & 0.654 & 0.676 & 0.674 & 0.660 & 0.650 & --- \\
SA-AH-GRPO ($\alpha$=0.5)
  & 0.648 & 0.644 & 0.664 & 0.682 & 0.670 & 0.686 \\
\bottomrule
\end{tabular}
\end{table}

\paragraph{Accuracy (3B model).}
All three methods converge to similar final Pass@1 values in the range
$0.844$--$0.862$, well above the Qwen\,2.5-3B-Instruct zero-shot baseline of
0.831. The zero-shot baseline is already strong for this model, so the
primary benefit of RL fine-tuning at this scale is stability rather than
absolute accuracy gains.

\paragraph{Accuracy (1.5B model).}
Starting from a weaker zero-shot baseline of 0.637, all three methods improve
accuracy. SA-AH-GRPO achieves a peak and final Pass@1 of \textbf{0.686}, a
$+4.9$ percentage-point improvement over the baseline. Standard GRPO peaks at
0.668 and AH-GRPO ($\alpha{=}0.5$) at 0.676, showing that the asymmetric
discounting variant provides a consistent benefit at the smaller scale.

\paragraph{Training stability.}
Training variance is the variance of mean per-step reward over the second half of the training run. On the 3B model, SA-AH-GRPO achieves a training variance of
$0.0246$---a \textbf{3.6$\times$ reduction} relative to GRPO ($0.0885$) and a
\textbf{2.6$\times$ reduction} relative to AH-GRPO ($0.0630$). On the 1.5B
model, SA-AH-GRPO shows a higher training variance ($0.1718$) than the other
methods, consistent with a model that has more room to improve and thus
experiences greater reward variation during training.

\paragraph{KL divergence.}
Mean per-token KL divergence from the reference policy is low across all
methods ($<0.010$). On the 3B model, SA-AH-GRPO shows a slightly higher
mean KL ($0.00785$ vs.\ $0.00512$ for AH-GRPO), consistent with more active
exploration enabled by unrestricted gradient flow on positive-advantage
rollouts. The 1.5B SA-AH-GRPO run shows the highest mean KL ($0.00986$),
reflecting the greater policy update magnitude at that scale.

\subsection{$\alpha$-Ablation of AH-GRPO on Qwen\,2.5-1.5B-Instruct}
\label{sec:alpha_ablation}

To characterise the sensitivity of AH-GRPO to the entropy discount strength
$\alpha$, we conduct a sweep over $\alpha \in \{-0.25,\, 0.0,\, 0.10,\, 0.25,\, 0.50\}$
on the 1.5B model (150 steps each). Negative $\alpha$ inverts the discount,
\emph{up-weighting} high-entropy positions (an entropy-amplifying regime).
All runs use the same hyperparameters as the main experiments.

\begin{table}[H]
\centering
\caption{AH-GRPO $\alpha$-ablation on \textbf{Qwen\,2.5-1.5B-Instruct}
         (150 steps, greedy decoding, 500 examples).
         Pass@1 at each evaluated checkpoint; final and peak summarised.
         Partial data available for $\alpha{=}0.25$ (steps 30--60 only).
         $\uparrow$ = higher is better.}
\label{tab:alpha_ablation}
\setlength{\tabcolsep}{5pt}
\begin{tabular}{lccccccc}
\toprule
\multirow{2}{*}{$\alpha$} &
\multicolumn{5}{c}{Pass@1 by Step} &
\multirow{2}{*}{Peak $\uparrow$} &
\multirow{2}{*}{Final $\uparrow$} \\
\cmidrule(lr){2-6}
& 30 & 60 & 90 & 120 & 150 & & \\
\midrule
$-0.25$ & 0.648 & 0.632 & 0.630 & 0.650 & 0.634 & 0.650 & 0.634 \\
$\phantom{-}0.00$ (GRPO) & 0.656 & 0.666 & 0.650 & 0.668 & 0.660 & 0.668 & 0.660 \\
$\phantom{-}0.10$ & 0.670 & 0.658 & 0.664 & 0.664 & 0.640 & \textbf{0.670} & 0.640 \\
$\phantom{-}0.25$ & 0.676 & 0.654 & 0.650 & 0.648 & 0.654 & 0.676$^{\dagger}$ & 0.654 \\
$\phantom{-}0.50$ & 0.654 & 0.676 & 0.674 & 0.660 & 0.650 & 0.676 & 0.650 \\
\midrule
SA-AH-GRPO $0.50$ & 0.648 & 0.644 & 0.664 & 0.682 & 0.670 & --- & --- \\
\quad (180 steps) & --- & --- & --- & --- & --- & 0.686 & 0.686 \\
\bottomrule
\end{tabular}
\end{table}
\vspace{-4pt}
{\small $^\dagger$ Partial run (steps 30--60 only); peak may not be the global maximum.}

\paragraph{Discussion of ablation.}
Positive $\alpha$ values (entropy-discounting) consistently outperform the
negative setting ($\alpha = -0.25$), which degrades below baseline in later
training steps. This confirms the motivation for AH-GRPO: down-weighting
high-entropy token positions is beneficial, whereas amplifying them hurts.
Among positive values, $\alpha = 0.10$ achieves the highest peak Pass@1
(0.670) but the lowest final value (0.640), suggesting a trade-off between
early accuracy and late-stage stability at small $\alpha$. $\alpha = 0.50$
produces slightly lower peak but better final accuracy (0.650). The
SA-AH-GRPO result at the same $\alpha$ (0.686 peak/final) further
demonstrates that restricting the discount to negative-advantage rollouts
is the decisive improvement over symmetric AH-GRPO.

\subsection{Trajectory Analysis}
\label{sec:traj}

\begin{figure}[H]
\centering
\includegraphics[width=0.88\textwidth]{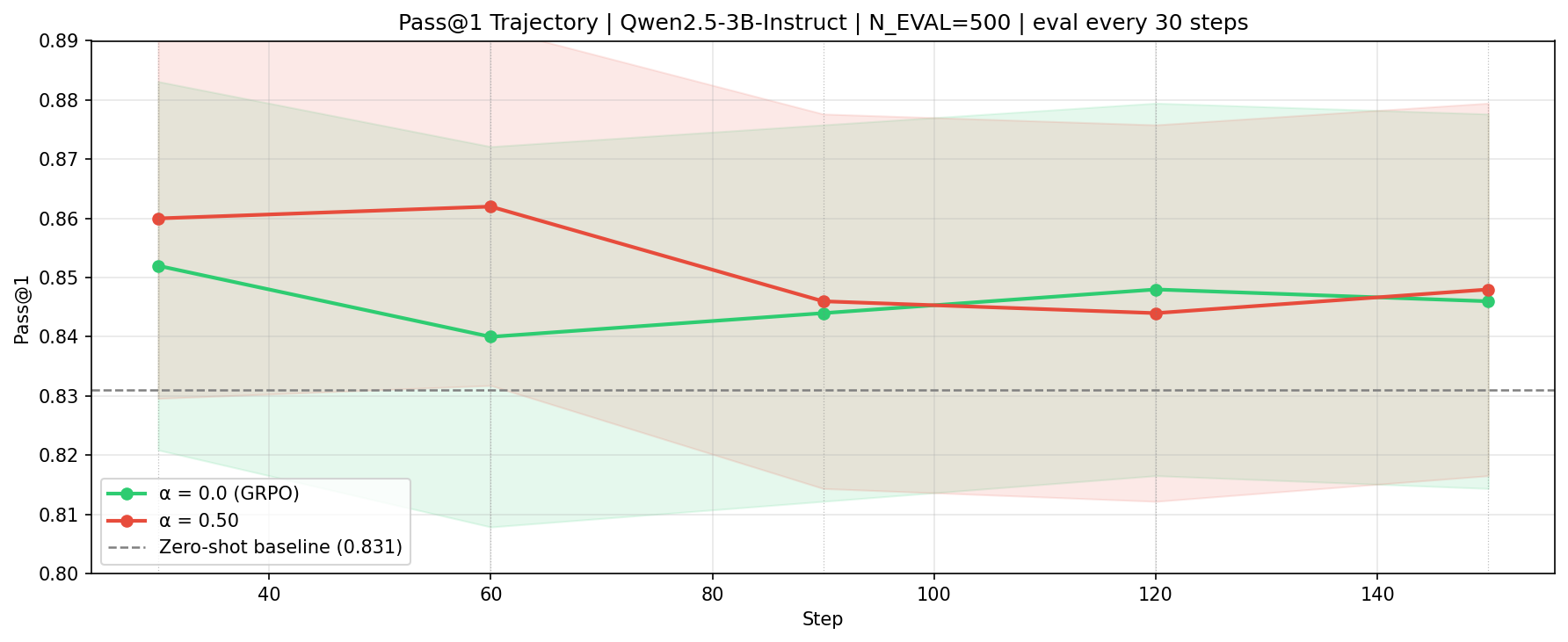}
\caption{Pass@1 on GSM8K test split (500 examples) vs.\ training step for
GRPO ($\alpha{=}0$) and AH-GRPO ($\alpha{=}0.5$) on
\textbf{Qwen\,2.5-3B-Instruct} (greedy decoding, $N_{\mathrm{eval}}{=}500$,
evaluated every 30 steps). Shaded bands show 95\% binomial confidence
intervals. Dashed line: zero-shot baseline (0.831).}
\label{fig:pass1_3b_alpha}
\end{figure}

\begin{figure}[H]
\centering
\includegraphics[width=0.88\textwidth]{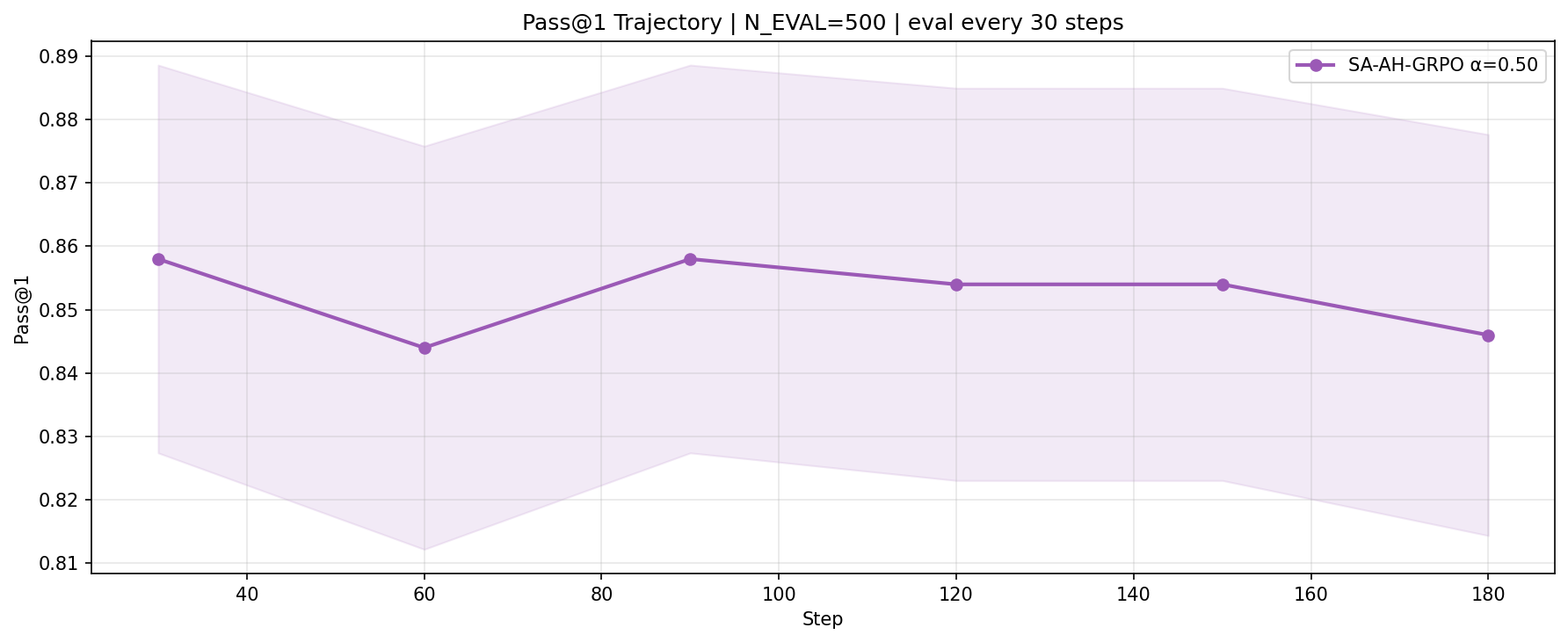}
\caption{Pass@1 on GSM8K test split (500 examples) vs.\ training step for
\textbf{SA-AH-GRPO ($\alpha{=}0.5$)} on \textbf{Qwen\,2.5-3B-Instruct}
(180 steps, evaluated every 30 steps, greedy decoding). Shaded band shows
95\% binomial confidence intervals. SA-AH-GRPO achieves a $\mathbf{3.6\times}$
reduction in training reward variance relative to GRPO (0.0246 vs.\ 0.0885;
see Table~\ref{tab:main}).}
\label{fig:pass1_3b}
\end{figure}

\begin{figure}[H]
\centering
\includegraphics[width=0.88\textwidth]{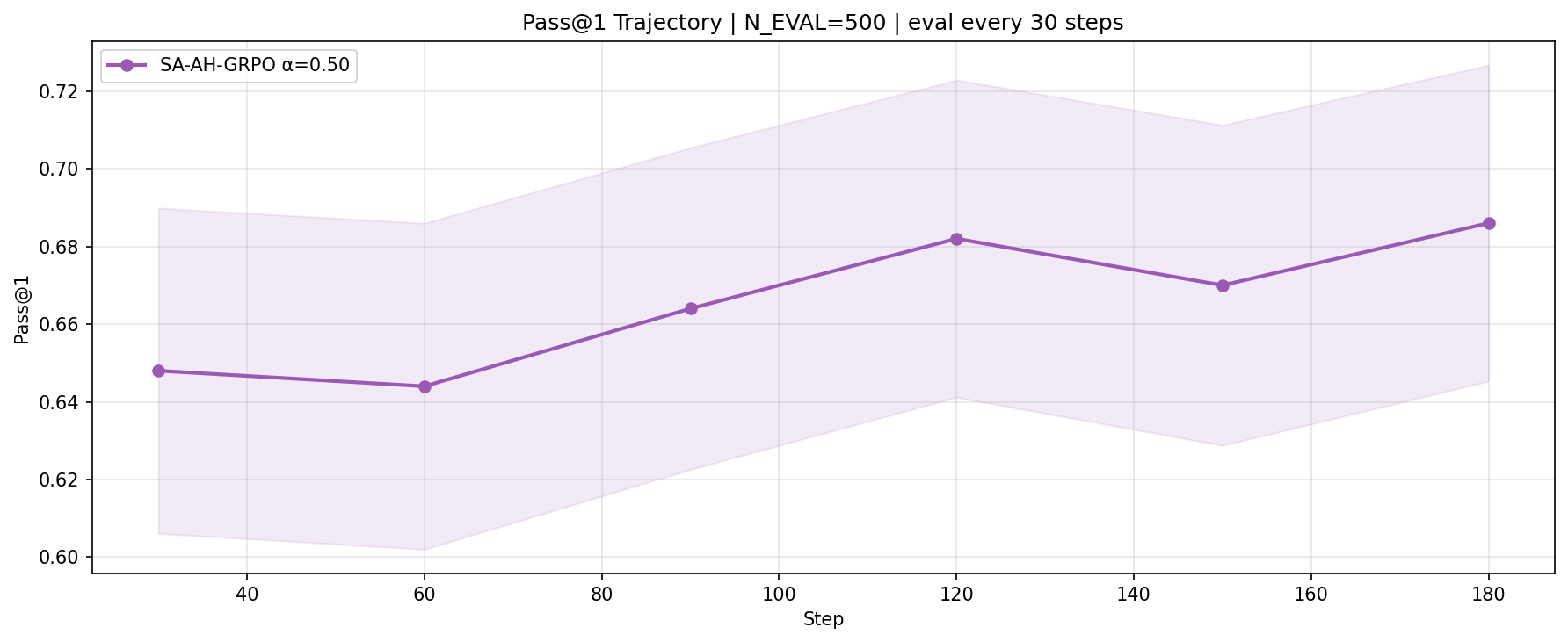}
\caption{Pass@1 on GSM8K test split (500 examples) vs.\ training step for
\textbf{SA-AH-GRPO ($\alpha{=}0.5$)} on \textbf{Qwen\,2.5-1.5B-Instruct}
(180 steps, evaluated every 30 steps, greedy decoding). Shaded band shows
95\% binomial confidence intervals. The model shows a consistent upward
trend, reaching a peak and final Pass@1 of \textbf{0.686} at step~180 ---
a $\mathbf{+4.9}$~pp improvement over the zero-shot baseline of 0.637.}
\label{fig:pass1_1b}
\end{figure}

\clearpage
\begin{figure}[H]
\centering
\includegraphics[width=0.72\textwidth]{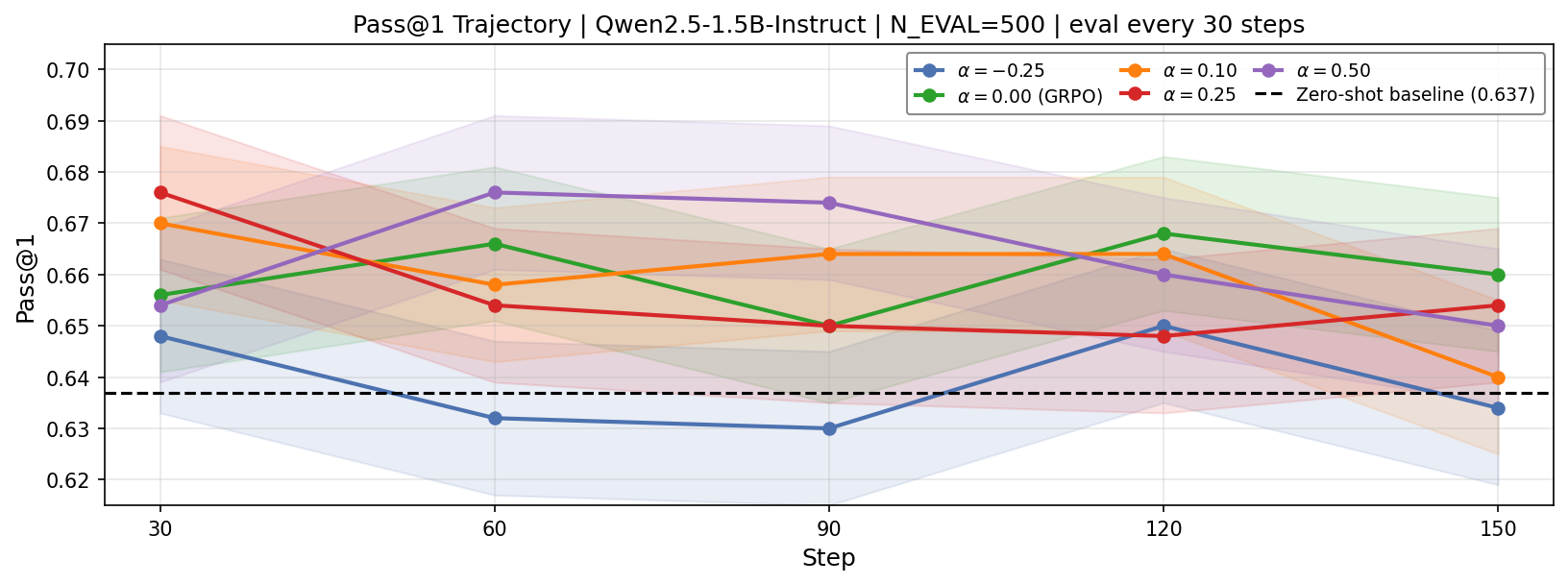}
\caption{$\alpha$-ablation of AH-GRPO on Qwen\,2.5-1.5B-Instruct (150~steps).
Solid circles: peak Pass@1. Open squares: final Pass@1 at step~150.
Dashed line: zero-shot baseline (0.637).
Positive $\alpha$ uniformly outperforms the entropy-amplifying regime
($\alpha{<}0$); $\alpha{=}0.10$ achieves the highest peak (0.670) while
$\alpha{=}0.50$ yields the best peak among the AH-GRPO runs
that also appear in the main comparison (0.676).}
\label{fig:alpha_ablation}
\end{figure}

Training logs from the 3B SA-AH-GRPO run reveal several notable patterns:

\begin{itemize}
    \item \textbf{Reward trajectory.} Mean group reward rises from
          $6.82 \pm 0.81$ at step 5 to $7.09 \pm 0.24$ at step 180, with
          low variance throughout later training (std $< 0.25$).
    \item \textbf{Negative-advantage fraction.} The fraction of rollouts with
          $\adv_i < 0$ (\texttt{neg\_frac}) fluctuates between 0 and 1 across
          steps, reflecting the stochastic nature of group normalisation.
          When \texttt{neg\_frac} = 0 (all rollouts are above-average),
          SA-AH-GRPO reduces to standard GRPO; when \texttt{neg\_frac} = 1,
          it applies the full AH discount.
    \item \textbf{Mean weight $\bar{w}$.} The average adaptive weight across
          all tokens and rollouts remains above 0.5 throughout training,
          indicating that the discount does not completely suppress gradients
          even on negative rollouts.
    \item \textbf{Response length.} Mean generation length decreases from
          $\sim$243 tokens at step 5 to $\sim$171 tokens at step 180,
          consistent with the model learning more concise, direct reasoning
          chains---a known side-effect of RLVR on format-rewarded tasks.
    \item \textbf{Entropy.} Mean normalised entropy $\bar{H}$ decreases
          from $\sim$0.029 at step 5 to $\sim$0.009 at step 180, indicating
          that the policy becomes increasingly confident over training. The
          AH discount therefore weakens over time, a natural form of annealing.
\end{itemize}

\section{Discussion}
\label{sec:discussion}

\paragraph{Why selective application stabilises training.}
Under standard GRPO and AH-GRPO, both correct and incorrect rollouts
contribute gradients that interact at every shared parameter. When a
high-entropy token position appears in a correct rollout, AH-GRPO discounts
its gradient even though that position legitimately participated in producing
the correct answer. This creates a systematic bias: the algorithm selectively
weakens evidence from the most uncertain---and often most informative---token
choices in successful solutions. SA-AH-GRPO eliminates this bias by
preserving the full gradient signal on positive-advantage rollouts,
resulting in less conflicting gradient directions and therefore lower
training variance (most clearly seen at the 3B scale).

\paragraph{Scale sensitivity.}
The variance reduction benefit of SA-AH-GRPO is most pronounced at the 3B
scale ($3.6\times$ vs.\ GRPO). At 1.5B, SA-AH-GRPO instead shows higher
training variance alongside better final accuracy, suggesting that at this
scale the model benefits more from the unrestricted gradient signal on
positive rollouts than from gradient dampening on negative ones. The 4.9
percentage-point gain over the zero-shot baseline ($0.637 \to 0.686$) for the
1.5B SA-AH-GRPO is consistent with a model that has more headroom to
improve and exploits the full positive-rollout gradient effectively.

\paragraph{Adaptive horizon as implicit curriculum.}
As the model trains and entropy decreases (Section~\ref{sec:traj}), the
per-token discount $\gamma_{t}^{(i)} = e^{-\alpha\hn_t}$ approaches 1, and the
effective GRPO and AH-GRPO losses converge. SA-AH-GRPO thus implements a
form of \emph{implicit curriculum}: early in training, when the model is
uncertain and negative rollouts have high entropy, the discount is strongest,
protecting the optimiser from noisy, conflicting gradients. Late in training,
the discount naturally anneals as the policy becomes more confident.

\paragraph{Comparison with related discounting approaches.}
Token-level weighting has been explored in various forms: advantage
regularisation in PPO \cite{schulman2017proximal}, per-token KL penalties
\cite{ziegler2019fine}, and length penalties \cite{yuan2024advancing}.
Our approach differs in that the weight is derived from the model's own
predictive uncertainty at each position, rather than from an external
heuristic, and is applied asymmetrically based on trajectory quality.

\section{Limitations}
\label{sec:limitations}

Our experiments are confined to two model sizes of the Qwen\,2.5 family and a
single benchmark (GSM8K). The $\alpha$-ablation is performed only for
AH-GRPO on the 1.5B model; a corresponding sweep for SA-AH-GRPO and for the
3B model remains future work, as does the $\alpha{=}0.25$ run which was
only partially completed. The top-$K$ entropy approximation ($K=500$)
introduces a minor inaccuracy in $\hn_t$; we did not ablate $K$. We also note
that the 95\% CI width ($\pm0.031$--$0.042$ depending on scale and accuracy)
means absolute Pass@1 differences of $<0.02$ should be interpreted cautiously.
All runs use a fixed random seed (123); variance across seeds is not reported
and may affect the generality of our stability conclusions.

\section{Related Work}
\label{sec:related}

\paragraph{RLHF and RLVR for language models.}
Reinforcement Learning from Human Feedback \cite{ouyang2022training} and
verifiable rewards \cite{lightman2023lets, cobbe2021gsm8k} have become
standard tools for aligning large language models. PPO \cite{schulman2017proximal}
remains the dominant algorithm, with GRPO \cite{shao2024deepseekmath}
offering a critic-free alternative particularly suited to reasoning tasks.

\paragraph{Policy gradient variance reduction.}
Variance in policy gradient methods has been extensively studied
\cite{williams1992simple, schulman2015high}. Advantage normalisation,
baselines, and actor-critic architectures are standard tools. GRPO's
group-relative baseline is an instance of the within-group baseline idea.
Our entropy-adaptive weighting provides an orthogonal source of variance
reduction at the token level.

\paragraph{Token-level reward shaping.}
Process reward models \cite{lightman2023lets} assign per-step (rather than
outcome) rewards, providing denser supervision. Dense token-level rewards
have also been explored in \cite{xu2024finegrained}. Our approach does not require
an auxiliary process reward model; instead, it modulates the \emph{weighting}
of the existing outcome reward signal.

\paragraph{Entropy regularisation.}
Maximum-entropy RL \cite{ziegler2019fine, haarnoja2018soft} encourages
exploration by adding an entropy bonus to the reward. Our use of entropy is
complementary: rather than maximising entropy, we use it as a signal to
\emph{down-weight} gradient contributions at uncertain positions, preventing
unstable updates.

\paragraph{Length and complexity control.}
Recent works have noted that RLVR can cause models to generate increasingly
long (or short) responses \cite{yuan2024advancing}. Our method implicitly
penalises long uncertain prefixes (the cumulative discount grows with
trajectory length under high entropy), which may partially explain the
observed reduction in response length.

\section{Conclusion}
\label{sec:conclusion}

We presented AH-GRPO and SA-AH-GRPO, two extensions of GRPO that introduce
entropy-adaptive token-level discounting for language model policy
optimisation. AH-GRPO applies the discount uniformly; SA-AH-GRPO restricts
it to negative-advantage rollouts, preserving the full gradient signal on
correct solutions. Experiments on GSM8K with Qwen\,2.5-3B-Instruct and
Qwen\,2.5-1.5B-Instruct show that SA-AH-GRPO achieves the same peak accuracy
as GRPO and AH-GRPO on the 3B model while reducing training variance by
$3.6\times$, and achieves the best final accuracy ($+4.9$ pp.\ over
zero-shot) on the 1.5B model. An $\alpha$-ablation of AH-GRPO on the 1.5B
model confirms that positive entropy-discounting is beneficial, while
entropy-amplifying ($\alpha < 0$) settings degrade performance. These results
suggest that asymmetric entropy discounting is a principled and practical
improvement to GRPO-based RLVR fine-tuning.

Future work will evaluate SA-AH-GRPO at larger model scales (7B, 70B),
across task domains (code synthesis, logical reasoning), and with a full
$\alpha$-sweep for SA-AH-GRPO on both model sizes to characterise the joint
sensitivity of asymmetry and discount strength.

\appendix
\section{Remaining Figures To Generate}
\label{app:diagrams}

Figures~\ref{fig:pass1_3b_alpha}, \ref{fig:pass1_3b}, and~\ref{fig:pass1_1b}
(Pass@1 trajectories) are included in the main paper from experimental outputs.
Figure~\ref{fig:alpha_ablation} ($\alpha$-ablation) is generated from the
re-eval logs. The following additional figures are recommended for
a camera-ready submission, all generatable from the training logs:

\begin{enumerate}[leftmargin=*]
  \item \textbf{Combined 3B Pass@1 overlay (all three methods).}
        GRPO, AH-GRPO, and SA-AH-GRPO on a single panel with
        a dashed zero-shot baseline at 0.831 and 95\% CI bands.
  \item \textbf{$\alpha$-ablation bar/line chart} (Fig.~\ref{fig:alpha_ablation}).
        Peak and final Pass@1 for AH-GRPO
        $\alpha \in \{-0.25, 0, 0.10, 0.25, 0.50\}$ on the 1.5B model.
        Generate from \texttt{ah\_grpo\_sweep/} re-eval logs.
  \item \textbf{Training reward mean $\pm$ std over steps.}
        Overlay all three 3B methods. Highlights SA-AH-GRPO variance
        reduction (Var $= 0.0246$ vs.\ $0.0885$ for GRPO).
  \item \textbf{Entropy $\bar{H}$ and adaptive weight $\bar{w}$ over steps.}
        SA-AH-GRPO 3B. Illustrates the implicit-curriculum / natural-annealing
        property as entropy decays from 0.029 to 0.009.
  \item \textbf{Schematic / method diagram.}
        Side-by-side illustration of token-weight assignment for GRPO,
        AH-GRPO, and SA-AH-GRPO on a positive vs.\ negative rollout.
        Makes the asymmetry of SA-AH-GRPO immediately intuitive.
\end{enumerate}

\section{Algorithm Pseudocode}

\begin{algorithm}[H]
\caption{SA-AH-GRPO Training Step}
\label{alg:saahgrpo}
\begin{algorithmic}[1]
\Require Policy $\pi_\theta$, reference $\pi_{\mathrm{ref}}$, prompt $q$,
         hyperparameters $G, \alpha, \beta, \epsilon$
\State Sample completions $\{o_i\}_{i=1}^{G} \sim \pi_{\theta_{\mathrm{old}}}(\cdot \mid q)$
\State Compute rewards $r_i = r(o_i, q, a^*)$ for each completion
\State Compute advantages $\adv_i = (r_i - \bar{r})/(\sigma_r + \varepsilon)$
\For{each rollout $i = 1,\ldots,G$}
  \For{each token position $t = 1,\ldots,|o_i|$}
    \State Compute logits $\bm{f}_t^{(i)} = \pi_\theta(\cdot \mid q, o_{i,<t})$
    \State $\hn_t^{(i)} \gets \frac{-\sum_{v\in\text{top-}K} p_{tv}\log p_{tv}}{\log V}$
           \Comment{normalised entropy, top-$K{=}500$}
    \State $\wt^{(i)} \gets \exp\!\left(-\alpha\sum_{s=1}^{t}\hn_s^{(i)}\right)$
           \Comment{cumulative AH weight (log-space)}
    \If{$\adv_i \geq 0$}  \Comment{positive rollout: no discount}
      \State $\tilde{w}_t^{(i)} \gets 1$
    \Else  \Comment{negative rollout: apply AH discount}
      \State $\tilde{w}_t^{(i)} \gets \wt^{(i)}$
    \EndIf
  \EndFor
\EndFor
\State Compute weighted clipped surrogate loss $\cL_{\mathrm{SA}}$ (Eq.~\ref{eq:saahgrpo})
\State Update $\theta \gets \theta - \eta \nabla_\theta \cL_{\mathrm{SA}}$
\end{algorithmic}
\end{algorithm}

\section{Detailed Training Logs}
\label{app:logs}

Tables~\ref{tab:sa_log_3b}--\ref{tab:sa_log_1b} report per-step diagnostic
metrics from the SA-AH-GRPO runs on both model scales (sampled every 30 steps
for brevity).

\begin{table}[H]
\centering
\caption{SA-AH-GRPO diagnostic metrics at selected training steps ---
         \textbf{Qwen\,2.5-3B-Instruct}.
         $\bar{r}$: mean group reward ($\pm$std); KL: mean per-token KL;
         $\bar{H}$: mean normalised entropy; $\bar{w}$: mean adaptive weight
         (all tokens); $\bar{w}_-$: mean weight on negative-rollout tokens
         ($^\dagger$0.000 = all rollouts positive at that step, no discount applied);
         $f_-$: fraction of rollouts with $\adv < 0$; len: mean completion length.}
\label{tab:sa_log_3b}
\setlength{\tabcolsep}{4.5pt}
\begin{tabular}{rcccccccc}
\toprule
Step & Pass@1 & $\bar{r}$ & KL & $\bar{H}$ & $\bar{w}$ & $\bar{w}_-$ & $f_-$ & len \\
\midrule
  5  & ---   & $6.815\pm0.809$ & 0.00373 & 0.0291 & 0.516 & 0.174 & 0.50 & 243 \\
 30  & 0.858 & $6.933\pm0.602$ & 0.00396 & 0.0232 & 1.000 & 0.000$^{\dagger}$ & 0.00 & 213 \\
 60  & 0.844 & $7.202\pm0.115$ & 0.00821 & 0.0114 & 1.000 & 0.000$^{\dagger}$ & 0.00 & 242 \\
 90  & 0.858 & $6.782\pm0.163$ & 0.00472 & 0.0246 & 1.000 & 0.000$^{\dagger}$ & 0.00 & 250 \\
120  & 0.854 & $7.040\pm0.488$ & 0.00542 & 0.0085 & 0.891 & 0.584 & 0.25 & 212 \\
150  & 0.854 & $6.820\pm0.197$ & 0.02236 & 0.0111 & 0.912 & 0.609 & 0.25 & 224 \\
180  & 0.846 & $7.085\pm0.240$ & 0.00894 & 0.0092 & 0.805 & 0.619 & 0.50 & 171 \\
\bottomrule
\end{tabular}
\end{table}

\begin{table}[H]
\centering
\caption{SA-AH-GRPO diagnostic metrics at selected training steps ---
         \textbf{Qwen\,2.5-1.5B-Instruct}.
         Column definitions as in Table~\ref{tab:sa_log_3b}.}
\label{tab:sa_log_1b}
\setlength{\tabcolsep}{4.5pt}
\begin{tabular}{rcccccccc}
\toprule
Step & Pass@1 & $\bar{r}$ & KL & $\bar{H}$ & $\bar{w}$ & $\bar{w}_-$ & $f_-$ & len \\
\midrule
  5  & ---   & $2.980\pm2.000$ & 0.00494 & 0.0432 & 0.580 & 0.143 & 0.50 & 151 \\
 30  & 0.648 & $3.767\pm2.090$ & 0.00374 & 0.0307 & 0.997 & 0.810 & 0.25 & 131 \\
 60  & 0.644 & $5.757\pm1.411$ & 0.00413 & 0.0195 & 0.618 & 0.462 & 0.75 & 186 \\
 90  & 0.664 & $5.505\pm1.529$ & 0.00756 & 0.0344 & 0.636 & 0.169 & 0.50 & 205 \\
120  & 0.682 & $6.350\pm1.571$ & 0.00703 & 0.0153 & 0.677 & 0.439 & 0.50 & 156 \\
150  & 0.670 & $6.125\pm1.040$ & 0.00949 & 0.0211 & 0.668 & 0.174 & 0.25 & 181 \\
180  & 0.686 & $7.205\pm0.255$ & 0.01630 & 0.0246 & 0.593 & 0.340 & 0.50 & 142 \\
\bottomrule
\end{tabular}
\end{table}


\end{document}